\documentclass[11pt]{article}
\usepackage{acl}
\usepackage{times}
\usepackage{latexsym}
\usepackage[T1]{fontenc}
\usepackage[utf8]{inputenc}
\usepackage{microtype}
\usepackage{graphicx}
\usepackage{booktabs}
\usepackage{amsmath}
\usepackage{amssymb}
\usepackage[table]{xcolor}
\usepackage{multirow}
\usepackage{enumitem}
\usepackage{tikz}
\usetikzlibrary{arrows.meta,positioning}
\usepackage[ruled,vlined]{algorithm2e}
\usepackage{tcolorbox}
\tcbuselibrary{skins,breakable}
\graphicspath{{figures/output/}}

\definecolor{passone}{RGB}{66, 133, 244}    
\definecolor{passtwo}{RGB}{234, 67, 53}     
\definecolor{passthree}{RGB}{52, 168, 83}   
\definecolor{highlight}{RGB}{255, 235, 238} 
\definecolor{bestrow}{RGB}{232, 245, 233}   
\definecolor{promptbg}{RGB}{248, 249, 250}  
\definecolor{promptborder}{RGB}{66, 133, 244} 
\definecolor{citationblue}{RGB}{0, 106, 166}
\newcommand{\rev}[1]{#1}
\newcommand{\kaonslink}{Kaons~\href{https://kaons.com}{\textcolor{citationblue}{$\boldsymbol{K}^{\ast}$}}}

\usepackage{pifont}
\newcommand{\cmark}{\ding{51}}
\newcommand{\xmark}{\ding{55}}
\newif\ifincludesupplement
\includesupplementtrue
\hypersetup{
  colorlinks=true,
  linkcolor=citationblue,
  citecolor=citationblue,
  urlcolor=citationblue,
  pdftitle={AsymVerify at SemEval-2026 Task 6: Asymmetric Confidence-Gated Verification for Political Evasion Detection},
  pdfauthor={Sebastien Kawada},
  pdfsubject={SemEval-2026 Task 6 CLARITY system description paper},
  pdfkeywords={SemEval-2026, CLARITY, political evasion detection, response clarity classification, AsymVerify, AsymVerify-ACL, Kaons, code release}
}
\SetKwSty{textbf}
\SetArgSty{textnormal}
\SetDataSty{textnormal}
\SetFuncSty{textnormal}

\newtcolorbox{promptbox}[1][]{
  colback=promptbg,
  colframe=promptborder,
  fonttitle=\bfseries\footnotesize,
  title=#1,
  breakable,
  boxrule=0.4pt,
  arc=3pt,
  left=5pt, right=5pt, top=3pt, bottom=3pt,
  fontupper=\footnotesize,
  before skip=6pt,
  after skip=6pt,
  toptitle=2pt, bottomtitle=2pt
}

\title{AsymVerify at SemEval-2026 Task 6: Asymmetric Confidence-Gated Verification for Political Evasion Detection}

\author{Sebastien Kawada \\
  \kaonslink \\
  Los Angeles, CA, USA \\
  \texttt{sebastien@kaons.com}}

\begin{document}
\maketitle

\begin{abstract}
\rev{Political evasion is difficult to detect because evasive answers often appear cooperative while avoiding concrete commitment. We present AsymVerify, a confidence-gated verification system for SemEval-2026 Task 6, a three-way classification of Clear Reply, Ambivalent, and Clear Non-Reply responses. AsymVerify scored 0.85 Macro F1 on the evaluation split ($D_{\text{eval}}$, $n{=}237$), placing \textbf{2nd out of 41 teams} on the official leaderboard. The system first classifies each question-answer pair, then selectively applies downgrade verification (CR/CNR $\rightarrow$ AMB) or upgrade verification (AMB $\rightarrow$ CR) to low-confidence predictions. Development analysis shows that errors concentrate at the Ambivalent boundary in both directions, motivating this asymmetric two-verifier design while confidence gating keeps additional inference cost low. On $D_{\text{dev}}$ ($n{=}308$), AsymVerify with GLM-4.7 gains +17.1 Macro F1 over single-pass classification at 1.48 calls/example, and the upgrade verifier alone improves every tested LLM backend on $D_{\text{dev}}$ by +6.8 to +15.2 Macro F1 over its single-pass baseline. Code is available at \url{https://github.com/kaons-research/AsymVerify-ACL}.}
\end{abstract}

\section{Introduction}

Viewers often struggle to detect political question dodges without explicit cues \citep{clementson2018effects,clementson2018deceptively}. Large language models may face a related failure mode: preference-optimized systems can favor agreeable or user-aligned interpretations over truthfulness \citep{sharma2024sycophancy}, the opposite of what evasion detection requires. Politicians exploit this gap with rhetorical strategies that technically engage questions while avoiding concrete commitment \citep{bull1993how}. Skilled evasion appears substantive, employing topic-adjacent responses that feel informative while committing to nothing.

SemEval-2026 Task 6 requires classifying political question-answer pairs into Clear Reply (CR), Ambivalent (AMB), or Clear Non-Reply (CNR) \citep{thomas2026semeval2026task6clarity}. The task builds on the QEvasion taxonomy and corpus \citep{thomas-etal-2024-never}, and the CLARITY release provides 3,448 training examples with 59\% Ambivalent responses. The dataset consists of English-language U.S. political interview transcripts.

We evaluate on three dataset splits: $D_{\text{train}}$ (3,448 examples) for prompt development, $D_{\text{dev}}$ (308) for threshold tuning and ablations, and $D_{\text{eval}}$ (237) for official placement. On $D_{\text{eval}}$, the GPT-5.2 AsymVerify submission scored 0.85 on CLARITY Subtask 1 and placed \textbf{2nd out of 41 teams}. On $D_{\text{dev}}$, model-family analyses examine the mechanism across multiple LLM backends. Confidence analysis shows a clear accuracy gap between high- and low-confidence predictions, which is the basis for selective verification. We further observe that CR and CNR, as opposite speech acts, are rarely confused directly (3\% of errors) and are instead typically misclassified as AMB. Because both verification directions correct through AMB, either pass alone \rev{targets} a large share of errors.

\section{Background}

\subsection{Task Definition}

The task defines three response clarity classes:

\begin{itemize}[nosep,leftmargin=*]
\item \textbf{Clear Reply (CR)} --- direct answer with specific commitment.

\item \textbf{Ambivalent (AMB)} --- evasive through vague language, topic shifts, or implicit answers.

\item \textbf{Clear Non-Reply (CNR)} --- explicit refusal or claim of ignorance.
\end{itemize}

Table~\ref{tab:dataset} shows dataset statistics. The dev set has higher Ambivalent proportion (67\%) than training (59\%), \rev{making minority-class Macro F1 harder.} Systems are evaluated by Macro F1, \rev{the unweighted mean $(\mathrm{F1}_{\mathrm{CR}}+\mathrm{F1}_{\mathrm{AMB}}+\mathrm{F1}_{\mathrm{CNR}})/3$,} a standard choice for imbalanced multi-class classification because each class contributes equally to the final score \citep{sokolova2009systematic}. We report accuracy as a secondary metric.

\begin{table}[t]
\centering
\normalsize
\begin{tabular}{lcccc}
\toprule
\textbf{Split} & \textbf{N} & \textbf{AMB} & \textbf{CR} & \textbf{CNR} \\
\midrule
Train & 3,448 & 59\% & 31\% & 10\% \\
Dev & 308 & 67\% & 26\% & 8\% \\
Eval & 237 & \multicolumn{3}{c}{\textit{(official placement)}} \\
\bottomrule
\end{tabular}
\caption{Dataset splits. Dev set skews toward Ambivalent (67\% vs 59\%).}
\label{tab:dataset}
\end{table}

\paragraph{Dataset notation.}
To avoid cross-split ambiguity, we use $D_{\text{train}}$ (n=3,448), $D_{\text{dev}}$ (n=308), and $D_{\text{eval}}$ (n=237). Unless stated otherwise, prompt tuning and ablations are reported on $D_{\text{dev}}$, while official placement claims use $D_{\text{eval}}$.

\subsection{Related Work}

\citet{bull1993how} typologize tactics for eluding questions in political interviews, while \citet{clayman2001answers} shows how interviewees can resist questions while preserving the norm of answering. \citet{bavelas1990equivocal} introduced the Situational Theory of Communicative Conflict, arguing that equivocation arises from communicative dilemmas where all direct responses carry negative consequences. Recent political-QA work measures answer quality through question-answer fit \citep{alvarez2026answerquality}. Our task differs from stance detection \citep{mohammad2016semeval}, which classifies opinion direction rather than response clarity, because evasion detection requires reasoning about whether commitments were made, not what position was taken.

Unlike self-consistency \citep{wang2023selfconsistency}, which improves accuracy by voting over multiple LLM samples, AsymVerify uses confidence to route examples selectively into verifier calls. This follows recent work on language-model cascades \citep{gupta2024lmcascades}, adaptive computation \citep{schuster2022calm}, and cost-aware LLM routing \citep{chen2023frugalgpt} that route inputs through different computational paths based on difficulty. Recent text-classification benchmarks also warn that heavier reasoning strategies can impose large token costs without uniform classification gains \citep{guo2026textreasoningbench}, so our verifier design applies extra reasoning only to uncertain boundary cases.

Our verification passes can be viewed as a task-specific form of iterative self-refinement \citep{madaan2023selfrefine,shinn2023reflexion}, where re-examination is asymmetrically conditioned on the initial prediction class rather than using open-ended self-correction, which can be unreliable without external feedback \citep{huang2024selfcorrect}. Structured reasoning has shown promise for implicit subjective classification \citep{sun2023textclassification,fei2023reasoning}, where surface-level features are insufficient. Evasion detection has the same property.

Verbalized confidence tends toward overconfidence \citep{groot2024overconfidence}, but recent work shows that prompted confidence can still provide useful uncertainty information when interpreted carefully \citep{xiong2024uncertainty,huang2024rankcalibration}. Newer uncertainty-estimation work distinguishes whether an uncertainty signal identifies hard cases from whether it is probabilistically calibrated \citep{li2025harmonized}, and AsymVerify uses confidence in the former sense, as a routing cue. RLHF can further degrade calibration, though verbalized confidence partially recovers it \citep{tian2023calibration}, and systematic sycophancy in RLHF-trained models causes them to prefer agreeable interpretations over accurate ones \citep{sharma2024sycophancy}. This is particularly problematic for evasion detection, where models must resist accepting vague answers as clear. Our prompt design \rev{counters} this tendency through structured taxonomy guidance and verification prompts that instruct skepticism.

\section{System Description}

AsymVerify operates in up to three passes, as shown in Figure~\ref{fig:control_flow}, with pseudocode in Appendix~\ref{app:algorithm}. The base classifier returns $(\hat{y}, c) = f_\theta(q, r)$ where $\hat{y} \in \{\textsc{cr}, \textsc{amb}, \textsc{cnr}\}$ and $c \in [0,1]$ is verbalized confidence, a self-reported scalar from structured JSON output rather than token logprobs \citep{tian2023calibration,xiong2024uncertainty}. High-confidence predictions ($c \geq \tau$) exit immediately, in the spirit of cascade methods that defer harder instances and resolve easier ones with fewer calls \citep{gupta2024lmcascades}. In the reported $D_{\text{dev}}$ analyses, low-confidence predictions keep the single Pass~1 label and receive targeted verification rather than maj@3 voting.

\begin{figure}[t]
\centering
\begingroup
\normalsize
\setlength{\abovedisplayskip}{2pt}
\setlength{\belowdisplayskip}{2pt}
\setlength{\abovedisplayshortskip}{2pt}
\setlength{\belowdisplayshortskip}{2pt}
\setlength{\fboxsep}{3pt}
\setlength{\jot}{1pt}
\newcommand{\stagebar}{\par\vspace{2pt}\noindent\rule{\linewidth}{0.35pt}\par\vspace{2pt}}
\newcommand{\routebox}[1]{\fcolorbox{black!30}{gray!8}{\strut\quad #1\quad}}
\begin{minipage}{\columnwidth}
\centering
{\bfseries Stage 1: Base Classification}\par\vspace{-2pt}
\[
\begin{aligned}
(\hat{y}_1,c_1)&=f_\theta(q,r),\quad \hat{y}_1\in\mathcal{Y}\\
\mathcal{Y}&=\{\mathrm{CR},\mathrm{AMB},\mathrm{CNR}\}
\end{aligned}
\]
\stagebar
{\bfseries Stage 2: Confidence Gate (single candidate)}\par\vspace{-2pt}
\[
\begin{aligned}
\hat{y} &= \hat{y}_1 && (c_1 \geq \tau;\ \mathrm{exit}),\\
\tilde{y} &= \hat{y}_1
&& (c_1 < \tau;\ \mathrm{verify}).
\end{aligned}
\]
\footnotesize Budget shown: single-candidate routing.
\stagebar
{\bfseries Stage 3: Conditional Verification ($c_1<\tau$)}\par\vspace{-2pt}
\[
\begin{aligned}
\mathcal{E} &= \{\mathrm{CR},\mathrm{CNR}\},\\
y^\downarrow &=
\begin{cases}
\mathrm{AMB}, & \tilde{y}\in\mathcal{E},\ g_\downarrow(q,r,\tilde{y})=1,\\
\tilde{y}, & \text{otherwise},
\end{cases}\\[1mm]
\hat{y} &=
\begin{cases}
\mathrm{CR}, & y^\downarrow=\mathrm{AMB},\ g_\uparrow(q,r)=1,\\
y^\downarrow, & \text{otherwise}.
\end{cases}
\end{aligned}
\]
\stagebar
{\bfseries Routing profile}\par\vspace{1mm}
\begin{tabular}{ccc}
\routebox{52.6\%} &
\routebox{29} &
\routebox{120} \\
early exit & P2 calls & P3 calls
\end{tabular}
\vspace{1mm}

\routebox{1.48 calls/example}
\end{minipage}
\endgroup
\caption{AsymVerify control flow. The $D_{\text{dev}}$ call budget uses single-candidate routing rather than maj@3. P2 can downgrade CR/CNR predictions to AMB before P3 optionally upgrades AMB predictions to CR. Routing counts are from GLM-4.7 on $D_{\text{dev}}$ ($n{=}308$; Table~\ref{tab:cost}).}
\label{fig:control_flow}
\end{figure}

The base classifier prompts the selected model with a structured evasion taxonomy containing nine response subtypes (explicit, implicit, general, partial, dodging, deflection, declining, claims ignorance, clarification). The official submission used GPT-5.2, while development analysis uses GLM-4.7, DeepSeek-V3.2, and Llama-3.3-70B on $D_{\text{dev}}$. The prompt emphasizes concrete commitments and instructs skepticism toward answers that sound substantive but avoid the specific question. The model returns structured JSON with label, confidence $c \in [0,1]$, and reasoning. All reported $D_{\text{dev}}$ call budgets use the single-candidate pipeline: each example receives one base classification call, and low-confidence examples are sent directly to the conditional verifiers. A maj@3 variant can replace the Pass~1 label, but it is not the cost setting reported in Table~\ref{tab:cost}.

Pass 2 re-examines CR and CNR predictions for possible downgrade to AMB using a ``one vs.\ multiple interpretations'' criterion, where if reasonable readers could disagree about what was actually said, the response is Ambivalent. Pragmatic accounts \citep{grice1975logic,bavelas1990equivocal,clayman2001answers} treat indirect answers as satisfying conversational relevance without making an explicit commitment. Pass 3 re-examines AMB predictions for possible upgrade to CR by checking whether the first substantive sentence directly answers the question while ignoring preambles and later tangents, a heuristic motivated by the turn-by-turn structure of news interviews \citep{clayman2002news,bull1993how}. We do not include an AMB$\rightarrow$CNR upgrade pass because CNR comprises only 8\% of the dataset and CNR$\rightarrow$AMB errors account for just \rev{5\%} of total errors (Table~\ref{tab:errors}), offering minimal recovery potential. Both verification passes run only on low-confidence predictions, and \rev{P3 runs on any low-confidence example whose current label is AMB after P2, which explains why verifier calls can slightly exceed the number of low-confidence examples.}

\paragraph{Prefilter for upgrade candidates.} Pass 3 can be expensive because 67\% of $D_{\text{dev}}$ examples are AMB. We therefore analyze, but do not use in the main GLM ablation, a rule-based prefilter that selects only AMB predictions whose first sentence shows strong commitment signals before sending them to the LLM verifier. Four lexical rules trigger verification: (1) answers starting with ``No'' followed by a short declarative sentence, (2) answers starting with ``Because'' (direct causal explanation), (3) answers starting with ``That is'' or ``That's'' (declarative assertion), and (4) the pattern ``No, I don't see'' (stance-taking). Applied to the GLM-4.7 full-system decisions, this prefilter reduced P3 calls from 120 to 3 and false upgrades from 19 to 0 (Table~\ref{tab:prefilter}).

The class error distribution further constrains verification design. CR and CNR are opposite speech acts that are rarely confused directly (3\% of errors) and are instead typically misclassified as AMB, so all corrections route through AMB.

\section{Experimental Setup}

\paragraph{Data.} We use $D_{\text{train}}$ (3,448) for prompt development, $D_{\text{dev}}$ (308) for threshold tuning and ablations, and $D_{\text{eval}}$ (237) for official placement. During submission, $D_{\text{eval}}$ labels were hidden, so model selection and analysis used $D_{\text{dev}}$.

\paragraph{Models.} The official submission used GPT-5.2 with \texttt{reasoning\_effort=high} \citep{openai2025gpt52}. Development-set ablations use GLM-4.7 as the primary analysis model \citep{openrouter2026glm47}, and model-family replication evaluates DeepSeek-V3.2 \citep{deepseek2025v32} and Llama-3.3-70B \citep{meta2024llama33} on $D_{\text{dev}}$. We report GPT-5.2 for the official evaluation split and use the development models for pass-level analysis, reflecting recent findings that model family, prompting strategy, and inference cost materially affect LLM text-classification behavior \citep{kostina2025classification,guo2026textreasoningbench}.

\paragraph{Hyperparameters.} Confidence threshold $\tau = 0.95$ selected from $\{0.85, 0.90, 0.95, 1.0\}$ on $D_{\text{dev}}$. Temperature is 0.1 for base classification and 0.0 for verification. Because LLM confidence measures can be miscalibrated even when they are useful for ranking difficulty \citep{groot2024overconfidence,huang2024rankcalibration,tian2023calibration,li2025harmonized}, we treat $\tau$ as a routing threshold rather than a calibrated probability. Each configuration uses one low-temperature decoding pass, and we add 1,000-sample paired bootstrap intervals for the close deltas most likely to be sampling-sensitive \citep{koehn2004statistical,dror2018hitchhiker}.

\section{Results}

We first report official leaderboard performance on $D_{\text{eval}}$ for CLARITY Subtask 1 from the shared-task overview \citep{thomas2026semeval2026task6clarity}, then analyze the mechanism on $D_{\text{dev}}$. \rev{Table~\ref{tab:results} gives the top-12 leaderboard.}

\begin{center}
\small
\setlength{\tabcolsep}{4pt}
\begin{tabular}{@{}r p{2.5cm} c@{}}
\toprule
\textbf{Rank} & \textbf{Participant} & \textbf{Prediction score} \\
\midrule
1 & TeleAI & \textbf{0.89} \\
\textbf{2} & \textbf{AsymVerify} & \textbf{0.85} \\
3 & CSE-UOI & 0.85 \\
4 & Rasende Rakete & 0.83 \\
5 & Evaluators & 0.83 \\
6 & YNU-HPCC & 0.83 \\
7 & moswisarut & 0.82 \\
8 & tahamunawar & 0.81 \\
9 & CLaC @ CLARITY & 0.80 \\
10 & SpinDetector & 0.80 \\
11 & gabriel\_stefan & 0.80 \\
12 & AGAI & 0.79 \\
\bottomrule
\end{tabular}
\captionof{table}{Official CLARITY Subtask 1 leaderboard on $D_{\text{eval}}$ (n=237; 41 teams). Scores are official prediction scores, and AsymVerify placed \textbf{2nd out of 41 teams}, with CSE-UOI also displaying 0.85.}
\label{tab:results}
\end{center}

\subsection{Development Analysis (Dev Set)}

Table~\ref{tab:portability} tests whether the verification mechanism is tied to a single backend. The P3 upgrade branch improves Macro F1 by +6.8 to +15.2 across all three $D_{\text{dev}}$ backends (GLM-4.7, DeepSeek-V3.2, Llama-3.3-70B). P2 is more model-sensitive: it gives the best GLM-4.7 result but over-corrects DeepSeek and Llama, indicating that downgrade verification needs model-specific thresholding.

\begin{table}[t]
\centering
\footnotesize
\setlength{\tabcolsep}{3pt}
\begin{tabular}{@{}llccc@{}}
\toprule
\textbf{Model} & \textbf{Config.} & \textbf{F1} & \textbf{Acc.} & \textbf{Calls/ex.} \\
\midrule
GLM-4.7 & P1 & 55.9\% & 77.6\% & 1.00 \\
GLM-4.7 & P1+P3 & 70.8\% & 72.1\% & 1.33 \\
GLM-4.7 & P1+P2+P3 & \textbf{73.0\%} & 75.3\% & 1.48 \\
DeepSeek-V3.2 & P1 & 55.9\% & 73.4\% & 1.00 \\
DeepSeek-V3.2 & P1+P3 & \textbf{62.7\%} & 72.1\% & 1.66 \\
DeepSeek-V3.2 & P1+P2+P3 & 55.3\% & 67.2\% & 1.70 \\
Llama-3.3-70B & P1 & 41.0\% & 70.8\% & 1.00 \\
Llama-3.3-70B & P1+P3 & \textbf{56.3\%} & 74.4\% & 1.94 \\
Llama-3.3-70B & P1+P2+P3 & 53.7\% & 73.1\% & 1.97 \\
\bottomrule
\end{tabular}
\caption{Model-family replication on $D_{\text{dev}}$ ($n{=}308$). P3-only improves every tested backend, while full P2+P3 is strongest on GLM-4.7 but not uniformly best.}
\label{tab:portability}
\end{table}

\subsection{Cross-Model Class Stability}
\label{sec:stability}

Improvements are not uniform across classes. Ambivalent detection remains stable across the best replicated variants (79.8--83.1 F1), suggesting that the evasion-focused prompt template transfers reliably even when the base model changes. In contrast, Clear Reply varies by 11.8 points (51.2--63.0), and Clear Non-Reply varies by 40.5 points (34.5--75.0), indicating that explicit refusal detection is substantially more model-dependent.

This asymmetry explains why portability gains can be large in Macro F1 while overall accuracy remains in a narrow band (72.1--75.3\%). Models can converge on majority-class AMB behavior yet still diverge on minority classes, especially CNR. Practically, \rev{backend replacement appears less sensitive for AMB-heavy analyses, while CNR-sensitive use cases require model-specific prompt adaptation and threshold re-tuning.}

\rev{The full confusion matrix in Table~\ref{tab:confusion} provides the bridge from class-level stability to error mechanism: the largest off-diagonal cells are AMB$\rightarrow$CR and CR$\rightarrow$AMB, while direct CR$\leftrightarrow$CNR swaps are sparse.} The aggregated error distribution in Table~\ref{tab:errors} makes the pattern explicit. The two dominant errors are AMB$\rightarrow$CR (55\%), where evasive responses are over-credited as commitments, and CR$\rightarrow$AMB (29\%), where clear commitments are penalized for rhetorical hedging. Direct CR$\leftrightarrow$CNR confusion is rare (3\%), showing that errors concentrate at the boundaries with AMB rather than between the polar classes.

\begin{table}[t]
\centering
\small
\setlength{\tabcolsep}{5pt}
\begin{tabular}{lccc}
\toprule
\textbf{Gold} & \textbf{Pred CR} & \textbf{Pred AMB} & \textbf{Pred CNR} \\
\midrule
CR & 56 & 22 & 1 \\
AMB & 42 & 158 & 6 \\
CNR & 1 & 4 & 18 \\
\bottomrule
\end{tabular}
\caption{Confusion matrix on $D_{\text{dev}}$ ($n{=}308$), GLM-4.7 P1+P2+P3.}
\label{tab:confusion}
\end{table}

\begin{table}[t]
\centering
\small
\begin{tabular}{lcc}
\toprule
\textbf{Error Type} & \textbf{Count} & \textbf{\%} \\
\midrule
AMB $\rightarrow$ CR & 42 & 55\% \\
CR $\rightarrow$ AMB & 22 & 29\% \\
AMB $\rightarrow$ CNR & 6 & 8\% \\
CNR $\rightarrow$ AMB & 4 & 5\% \\
CR $\leftrightarrow$ CNR & 2 & 3\% \\
\bottomrule
\end{tabular}
\caption{Error patterns on $D_{\text{dev}}$ ($n{=}308$), GLM-4.7 P1+P2+P3. Direct CR$\leftrightarrow$CNR confusion is rare because errors concentrate at class boundaries.}
\label{tab:errors}
\end{table}

\subsection{Error Coverage by Verification Direction}

This boundary concentration has a practical consequence. Over 90\% of remaining errors fall along the two verification directions \rev{($E_{\downarrow}$ for AMB$\rightarrow$\{CR,CNR\} and $E_{\uparrow}$ for CR$\rightarrow$AMB, with arrows denoting gold$\rightarrow$predicted labels)}.
Table~\ref{tab:coverage} reorganizes the same 76 errors by \emph{which pass direction they align with}. Pass 2 (downgrade) \rev{aligns with} over-acceptance errors where AMB is predicted as CR/CNR (48 errors, 63.2\%). Pass 3 (upgrade) \rev{aligns with} under-recognized commitments where CR is predicted as AMB (22 errors, 28.9\%). Only 6 errors (7.9\%) fall outside both routes.

This decomposition explains why single-pass variants already recover most of the available gain. Both passes target large but complementary slices of boundary-concentrated errors, so either direction alone improves strongly (Table~\ref{tab:ablations}), while combining both gives a smaller but consistent additional gain.

\begin{table}[t]
\centering
\small
\begin{tabular}{lcc}
\toprule
\textbf{Error Family} & \textbf{Count} & \textbf{Share} \\
\midrule
P2-aligned (AMB$\rightarrow$CR/CNR) & 48 & 63.2\% \\
P3-aligned (CR$\rightarrow$AMB) & 22 & 28.9\% \\
Outside current passes & 6 & 7.9\% \\
\bottomrule
\end{tabular}
\caption{Remaining-error alignment by verification direction on $D_{\text{dev}}$ ($n{=}308$), derived from Table~\ref{tab:errors} (76 total errors).}
\label{tab:coverage}
\end{table}

The ablation in Table~\ref{tab:ablations} quantifies each pass's contribution on GLM-4.7. The gap between base accuracy (77.6\%) and Macro F1 (55.9\%) reflects the 67\% AMB class imbalance, where high accuracy is achievable by defaulting to the majority class while Macro F1 demands balanced performance. Each verification pass independently contributes about +15 points, and combining both yields +17.1. Selective verification achieves this at roughly half the call budget of running all three passes unconditionally (457 versus 924 calls).

\begin{center}
\footnotesize
\setlength{\tabcolsep}{3pt}
\begin{tabular}{lccccc}
\toprule
\textbf{Config.} & \textbf{F1} & \textbf{Acc.} & \textbf{Calls/ex.} & \textbf{P2} & \textbf{P3} \\
\midrule
P1 & 55.9\% & 77.6\% & 1.00 & 0 & 0 \\
P1+P2 & 70.9\% & 76.0\% & 1.15 & 45 & 0 \\
P1+P3 & 70.8\% & 72.1\% & 1.33 & 0 & 102 \\
P1+P2+P3 & \textbf{73.0\%} & 75.3\% & 1.48 & 29 & 120 \\
\bottomrule
\end{tabular}
\captionof{table}{Verification pass ablation on $D_{\text{dev}}$ ($n{=}308$), GLM-4.7. P2/P3 columns report activation counts.}
\label{tab:ablations}
\end{center}

Where do the computational savings come from? In the single-candidate GLM-4.7 full system, confidence gating routes 162 of 308 examples (52.6\%) directly to output after Pass~1, leaving 146 low-confidence examples. These examples trigger 149 verifier calls because three P2 downgrades to AMB are subsequently checked by P3. The full system uses \rev{$C_{\text{Asym}}=N+n_{P2}+n_{P3}=457$} total calls versus \rev{$C_{\text{all}}=3N=924$} for running all three passes unconditionally, a 50.5\% reduction. \rev{A low-confidence maj@3 variant would add two base calls per low-confidence example, $C_{\text{maj@3}}=C_{\text{Asym}}+2n_{\text{low}}=749$.}

\begin{center}
\small
\begin{tabular}{lrc}
\toprule
\textbf{Stage} & \textbf{Calls} & \textbf{\% of full} \\
\midrule
Pass 1 (all examples) & 308 & 33.3\% \\
Pass 2 + Pass 3 (low-conf only) & 149 & 16.1\% \\
\midrule
\textbf{AsymVerify total} & \textbf{457} & \textbf{49.5\%} \\
All 3 passes unconditionally & 924 & 100\% \\
\bottomrule
\end{tabular}
\captionof{table}{\rev{API call budget for single-candidate routing} on $D_{\text{dev}}$ (GLM-4.7, $n{=}308$, without maj@3 voting). Confidence gating reduces total calls by 50.5\% versus running all passes on every example. A low-confidence maj@3 variant would add two base calls for each of the 146 low-confidence examples, increasing the total from 457 to 749 calls, or from 1.48 to 2.43 calls/example.}
\label{tab:cost}
\end{center}

The routing threshold is supported by a simple confidence-bin check. In the GLM-4.7 full system, predictions at or above $\tau{=}0.95$ account for \rev{162 examples, exactly matching the early-exit count}, and are more accurate than the aggregate below-threshold set: 80.9\% versus 76.0\% (Appendix Table~\ref{tab:confidence}). The tiny 0.70--0.80 bin is not meaningful on its own, and these bins are not a full rank- or probability-calibration study \citep{huang2024rankcalibration,li2025harmonized}. Instead, the aggregate trend supports verbalized confidence as a routing signal for identifying harder examples.

\rev{Appendix~\ref{app:routing-diagnostics} also reports a Pass~3 prefilter analysis. The lexical filter reduces P3 calls from 120 to 3 and eliminates false upgrades (19 to 0), raising Macro F1 from 73.0 to 74.7, and because the rules are English lexical cues, we present this as an efficiency variant rather than the main system.}

Paired bootstrap intervals show that close ablation differences should not be over-read, following standard NLP practice for comparing systems on the same examples \citep{koehn2004statistical,dror2018hitchhiker}. P1+P2 and P1+P3 differ by only -0.1 Macro F1 points, with a 95\% CI of [-5.9, 5.9]. The prefilter delta is +1.7 points but its CI includes zero [-0.9, 4.1]. In contrast, Llama's P1$\rightarrow$P1+P3 gain is +15.2 points with CI [7.5, 23.1], a robust replication of the upgrade branch.

Appendix~\ref{app:examples} illustrates verification in action with representative success and failure cases from $D_{\text{dev}}$.

\section{Discussion}

Confidence routing works because high-confidence predictions are usually correct, while low-confidence predictions benefit from targeted verification. Additional calls on easy cases can add noise rather than signal, so confidence acts as a practical difficulty signal that concentrates computation where it matters \citep{gupta2024lmcascades,schuster2022calm}.

Different verification strategies converge because errors concentrate at class boundaries rather than spanning opposite classes (Section~\ref{sec:stability}). Both downgrade and upgrade verification correct through AMB, and since over-acceptance and under-recognition each account for many errors, either path alone addresses a substantial fraction. The rare CR$\leftrightarrow$CNR confusions (3\%) fall outside both paths but are too infrequent to dominate system behavior.

Residual errors cluster in hedged commitments, procedural refusals, and conditional commitments. These cases are difficult even for humans: QEvasion has Fleiss $\kappa = 0.644$ inter-annotator agreement \citep{thomas-etal-2024-never}, and political responses are often crafted to appear both decisive and non-committal \citep{clayman2001answers,clayman2002news,clementson2018deceptively}. Appendix~\ref{app:embeddings} further shows that semantic embeddings alone provide negligible class separation (silhouette = 0.001), so retrieval and similarity methods are unlikely to help, making prompt-based pragmatic reasoning the relevant alternative.

\rev{On $D_{\text{dev}}$, AsymVerify reaches 73.0 Macro F1 using 49.5\% of the unconditional three-pass call budget. The control flow is backend-agnostic, with verifier prompts tuned per model.} Majority-vote variants can be budgeted separately if used. Future gains will likely require richer discourse-level features or calibrated multi-model disagreement signals rather than additional majority voting alone.

\section{Conclusion}

AsymVerify \rev{scored 0.85 on CLARITY Subtask 1} ($D_{\text{eval}}$), placing \textbf{2nd out of 41 teams} on the official leaderboard. Its central result is that verification passes targeting opposite failure modes converge because both flow corrections through the middle Ambivalent class. Confidence gating exploits this convergence to reach 73.0 Macro F1 on GLM-4.7 $D_{\text{dev}}$ with 1.48 calls/example. The P3 upgrade branch alone replicates gains of +6.8 to +15.2 Macro F1 across GLM-4.7, DeepSeek-V3.2, and Llama-3.3-70B.

\section*{Limitations}

Our confidence threshold and verification prompts were optimized on English-language U.S. political interviews, so other political cultures, languages, or interview formats may require re-tuning. Political evasion exists on a spectrum, and QEvasion's $\kappa=0.644$ agreement means the system inherits judgment calls where ``correct'' classification can reflect annotator interpretation rather than ground truth. We also do not evaluate adversarially crafted evasions or noisy ASR transcripts.

The official score establishes performance of the GPT-5.2 submission, while pass-level analysis is reported for GLM-4.7, DeepSeek-V3.2, and Llama-3.3-70B on $D_{\text{dev}}$. Results use single low-temperature decoding per configuration, so close ablation differences should be read through the paired bootstrap intervals. Finally, each question-answer pair is classified independently, although multi-turn context, follow-ups, and speaker history may resolve cases that are ambiguous in isolation.

\section*{Ethics Statement}

AsymVerify is intended to support analysis of public political interviews by identifying whether a response gives a clear answer to the question asked. The labels describe response clarity rather than speaker intent or factual truth, which is important because equivocation can be strategically ambiguous rather than simply false \citep{bavelas1990equivocal,clayman2001answers,clementson2018effects}. System outputs should therefore be interpreted with conversational context and human judgment.

\clearpage
\bibliography{references}

\ifincludesupplement
\appendix
\onecolumn

\section{Full Pseudocode}
\label{app:algorithm}

\begin{algorithm}[h]
\small
\SetKwInOut{Input}{Input}
\SetKwInOut{Output}{Output}
\SetKw{And}{and}

\Input{Question $q$, response $r$}
\Output{Label $\hat{y} \in \{\textsc{cr}, \textsc{amb}, \textsc{cnr}\}$}
\textbf{Pass 1: Base classification}\;
$(\hat{y}, c) \gets f_\theta(q, r)$\;
\textbf{Confidence gating}\;
\If{$c \geq \tau$}{
    \Return $\hat{y}$\;
}
\textbf{Low confidence: single-candidate verifier routing}\;
$\tilde{y} \gets \hat{y}$\;
\textbf{Conditional branch P2: Downgrade CR/CNR $\to$ AMB}\;
\If{$\tilde{y} \in \{\textsc{cr}, \textsc{cnr}\}$}{
    \If{$g_\downarrow(q, r, \tilde{y}) = 1$}{
        $\tilde{y} \gets \textsc{amb}$\;
    }
}
\textbf{Conditional branch P3: Upgrade AMB $\to$ CR}\;
\If{$\tilde{y} = \textsc{amb}$}{
    \If{$g_\uparrow(q, r) = 1$}{
        $\tilde{y} \gets \textsc{cr}$\;
    }
}
$\hat{y} \gets \tilde{y}$\;
\Return $\hat{y}$\;

\caption{\textsc{AsymVerify}. Parameters: $\tau{=}0.95$. The $D_{\text{dev}}$ call budget uses single-candidate routing, with maj@3 treated as a separate variant. P2 and P3 are conditional branches rather than unconditional sequential calls, and $g_\downarrow, g_\uparrow \in \{0,1\}$ are LLM verifiers.}
\label{alg:asymverify}
\end{algorithm}

\subsection{Additional Routing Analysis}
\label{app:routing-diagnostics}

\rev{Tables~\ref{tab:confidence} and~\ref{tab:prefilter} report the confidence-bin and prefilter analyses that support the routing decisions summarized in the main results.}

\begin{table}[h]
\centering
\small
\setlength{\tabcolsep}{6pt}
\begin{tabular}{lrrr}
\toprule
\textbf{Confidence} & \textbf{n} & \textbf{Acc.} & \textbf{Err.} \\
\midrule
<0.70 & 16 & 62.5\% & 37.5\% \\
0.70--0.80 & 2 & 50.0\% & 50.0\% \\
0.80--0.90 & 34 & 76.5\% & 23.5\% \\
0.90--0.95 & 94 & 78.7\% & 21.3\% \\
$\geq$0.95 & 162 & 80.9\% & 19.1\% \\
\bottomrule
\end{tabular}
\caption{\rev{Verbalized-confidence bin accuracy} on $D_{\text{dev}}$ ($n{=}308$), GLM-4.7 full system.}
\label{tab:confidence}
\end{table}

\begin{table}[h]
\centering
\small
\setlength{\tabcolsep}{6pt}
\begin{tabular}{lrrrr}
\toprule
\textbf{Variant} & \textbf{F1} & \textbf{Acc.} & \textbf{P3 calls} & \textbf{False up.} \\
\midrule
No prefilter & 73.0\% & 75.3\% & 120 & 19 \\
Lexical prefilter & 74.7\% & 78.9\% & 3 & 0 \\
\bottomrule
\end{tabular}
\caption{\rev{Pass 3 prefilter analysis} on $D_{\text{dev}}$ ($n{=}308$), GLM-4.7 full system.}
\label{tab:prefilter}
\end{table}

\section{Prompt Templates}
\label{app:prompts}

This appendix reproduces the full prompts used in all three passes.

\subsection{Base Classification Prompt}

\begin{promptbox}[Pass 1: Base Classification]
\textbf{Role:} You are classifying political Q\&A exchanges for evasion.\smallskip

\textbf{Guidance}
\begin{itemize}[leftmargin=1.5em,topsep=1pt,itemsep=0pt]
\item Politicians are skilled at appearing to answer while actually evading.
\item Be skeptical of surface-level cooperation; look for concrete commitments.
\end{itemize}\smallskip

\textbf{Evasion taxonomy}\quad
\textit{Note:} Answers begin with speaker identification (e.g., ``President Trump.''). This is transcript formatting; ignore it.\smallskip

\textbf{Clear Reply (Explicit):} Direct answers providing specific information, a clear yes/no with commitment, or concrete numbers, names, dates, and policies.\smallskip

\textbf{Ambivalent:} Evasive responses:
\begin{enumerate}[leftmargin=1.7em,topsep=1pt,itemsep=0pt]
\item Implicit: hints without stating explicitly.
\item General: too vague, lacks specificity.
\item Partial: addresses only part of the question.
\item Dodging: ignores the question or changes topic.
\item Deflection: starts on topic but pivots away.
\end{enumerate}\smallskip

\textbf{Clear Non-Reply:} Explicit refusal:
\begin{enumerate}[leftmargin=1.7em,topsep=1pt,itemsep=0pt]
\item Declining: explicitly refuses (``I won't comment'').
\item Claims ignorance: says they do not know.
\item Clarification: asks for clarification instead.
\end{enumerate}\smallskip

\textbf{Task}\quad Analyze the exchange:\quad
\textbf{Question:} ``\{question\}''\quad \textbf{Answer:} ``\{answer\}''\smallskip

Check:
\begin{enumerate}[leftmargin=1.7em,topsep=1pt,itemsep=0pt]
\item What specific information is the question asking for?
\item Does the answer provide that specific information?
\item Is there evasion, deflection, or vagueness?
\end{enumerate}\smallskip

Output JSON:\quad \texttt{\{"classification": "Clear Reply" | "Ambivalent" | "Clear Non-Reply", "confidence": 0.0--1.0, "reasoning": "brief"\}}
\end{promptbox}

\subsection{Downgrade Verification Prompts (Pass 2)}

\begin{tcolorbox}[colback=passtwo!5, colframe=passtwo, fonttitle=\bfseries\footnotesize, title=Pass 2a: Clear Reply $\rightarrow$ Ambivalent, breakable, boxrule=0.4pt, arc=3pt, left=5pt, right=5pt, top=3pt, bottom=3pt, fontupper=\footnotesize, before skip=6pt, after skip=6pt, toptitle=2pt, bottomtitle=2pt]
\textbf{Input}\quad Question: ``\{question\}''\quad Answer: ``\{answer\}''\quad
\textit{(Skip speaker ID; focus on substantive response.)}\smallskip

\textbf{Decision:} Does this answer admit only one interpretation or multiple?\smallskip

\textbf{Clear Reply:} only one interpretation is possible; the answer explicitly commits to a position, no inference needed.\quad
\textbf{Ambivalent:} multiple interpretations are possible; inference is required.\smallskip

\textbf{Examples}\\
Q: ``Have you seen my chocolates?'' A: ``The children were in your room this morning.''\\
$\rightarrow$ \textbf{Ambivalent} (implies the children took them, but does not explicitly say so)\\
Q: ``Have you seen my chocolates?'' A: ``Yes, they are in the kitchen.''\\
$\rightarrow$ \textbf{Clear Reply} (only one interpretation)\smallskip

\textbf{Output:}\quad \texttt{\{"classification": "Clear Reply" | "Ambivalent", "reasoning": "brief"\}}
\end{tcolorbox}

\begin{tcolorbox}[colback=passtwo!5, colframe=passtwo, fonttitle=\bfseries\footnotesize, title=Pass 2b: Clear Non-Reply $\rightarrow$ Ambivalent, breakable, boxrule=0.4pt, arc=3pt, left=5pt, right=5pt, top=3pt, bottom=3pt, fontupper=\footnotesize, before skip=6pt, after skip=6pt, toptitle=2pt, bottomtitle=2pt]
\textbf{Input}\quad Question: ``\{question\}''\quad Answer: ``\{answer\}''\quad
\textit{(Skip speaker ID; focus on substantive response.)}\smallskip

\textbf{Decision:} Is this a Clear Non-Reply or Ambivalent?\smallskip

\textbf{Clear Non-Reply:} openly refuses to share information. The refusal is explicit and unambiguous.
\begin{itemize}[leftmargin=1.5em,topsep=1pt,itemsep=0pt]
\item ``I don't know'' / ``I'm not aware'' (claims ignorance)
\item ``I won't comment'' / ``No comment'' (declines)
\item ``What do you mean?'' (asks for clarification)
\end{itemize}\smallskip

\textbf{Ambivalent:} provides a response but allows multiple interpretations.
\begin{itemize}[leftmargin=1.5em,topsep=1pt,itemsep=0pt]
\item Leverages the subject to pivot elsewhere (deflection)
\item Gives information that does not answer the question
\item Appears to engage but does not commit
\end{itemize}\smallskip

\textbf{Examples}\\
Q: ``Have you seen my chocolates?'' A: ``You should not keep chocolates all around the house.''\\
$\rightarrow$ \textbf{Ambivalent} (deflects; no information about seeing chocolates)\\
Q: ``Have you seen my chocolates?'' A: ``I don't know where they are.''\\
$\rightarrow$ \textbf{Clear Non-Reply} (explicit claim of ignorance)\smallskip

\textbf{Output:}\quad \texttt{\{"classification": "Clear Non-Reply" | "Ambivalent", "reasoning": "brief"\}}
\end{tcolorbox}

\subsection{Upgrade Verification Prompt (Pass 3)}

\begin{tcolorbox}[colback=passthree!5, colframe=passthree, fonttitle=\bfseries\footnotesize, title=Pass 3: Upgrade Verification, breakable, boxrule=0.4pt, arc=3pt, left=5pt, right=5pt, top=3pt, bottom=3pt, fontupper=\footnotesize, before skip=6pt, after skip=6pt, toptitle=2pt, bottomtitle=2pt]
\textbf{Input}\quad Question: ``\{question\}''\quad Answer: ``\{answer\}''\\
Currently classified as \textbf{Ambivalent}. Check if it should be \textbf{Clear Reply}.\quad
\textit{(Skip speaker ID; inspect first substantive sentence.)}\smallskip

\textbf{Upgrade to Clear Reply if the first substantive sentence:}
\begin{enumerate}[leftmargin=1.7em,topsep=1pt,itemsep=0pt]
\item Directly answers with yes/no, a specific stance, or a clear position.
\item Does not start with preambles (``Well\ldots'', ``Look\ldots'', ``Let me\ldots'').
\item Is not immediately followed by ``but'', ``however'', or ``although''.
\end{enumerate}\smallskip

\textbf{Important:} what comes \emph{after} the first substantive sentence does not matter.
The key test: can you extract one clear answer from the opening?\smallskip

\textbf{Clear Reply examples:}\quad
``No, I don't see a contradiction\ldots'' (clear stance)\quad
``That is one of the options\ldots'' (specific commitment)\quad
``Because it takes time\ldots'' (direct causal)\smallskip

\textbf{Stays Ambivalent:}\quad
``Well, I think\ldots'' (preamble)\quad
``It depends on\ldots'' (conditional)\quad
``I wouldn't say\ldots'' (negation without stance)\smallskip

\textbf{Output:}\quad \texttt{\{"classification": "Clear Reply" | "Ambivalent", "reasoning": "brief"\}}
\end{tcolorbox}

\vspace{1.5em}

\section{Extended Model-Family Comparison}
\label{app:models}

Table~\ref{tab:full_models} shows the complete model-family development results.

\begin{table}[ht]
\centering
\small
\setlength{\tabcolsep}{4pt}
\renewcommand{\arraystretch}{1.08}
\begin{tabular}{llccc}
\toprule
\textbf{Model} & \textbf{Config.} & \textbf{Acc.} & \textbf{Macro F1} & \textbf{Calls/ex.} \\
\midrule
GLM-4.7 & P1 & 77.6\% & 55.9\% & 1.00 \\
GLM-4.7 & P1+P2 & 76.0\% & 70.9\% & 1.15 \\
GLM-4.7 & P1+P3 & 72.1\% & 70.8\% & 1.33 \\
GLM-4.7 & P1+P2+P3 & 75.3\% & 73.0\% & 1.48 \\
DeepSeek-V3.2 & P1 & 73.4\% & 55.9\% & 1.00 \\
DeepSeek-V3.2 & P1+P3 & 72.1\% & 62.7\% & 1.66 \\
DeepSeek-V3.2 & P1+P2+P3 & 67.2\% & 55.3\% & 1.70 \\
Llama-3.3-70B & P1 & 70.8\% & 41.0\% & 1.00 \\
Llama-3.3-70B & P1+P3 & 74.4\% & 56.3\% & 1.94 \\
Llama-3.3-70B & P1+P2+P3 & 73.1\% & 53.7\% & 1.97 \\
\bottomrule
\end{tabular}
\caption{Extended model-family comparison on $D_{\text{dev}}$ ($n{=}308$).}
\label{tab:full_models}
\end{table}

\begin{table}[ht]
\centering
\small
\setlength{\tabcolsep}{5pt}
\begin{tabular}{llccc}
\toprule
\textbf{Model} & \textbf{Config.} & \textbf{CR F1} & \textbf{AMB F1} & \textbf{CNR F1} \\
\midrule
GLM-4.7 & P1+P2+P3 & 62.9\% & 81.0\% & 75.0\% \\
DeepSeek-V3.2 & P1+P3 & 57.0\% & 79.8\% & 51.4\% \\
Llama-3.3-70B & P1+P3 & 51.2\% & 83.1\% & 34.5\% \\
\bottomrule
\end{tabular}
\caption{\rev{Per-class F1 for the strongest development variant of each tested backend} on $D_{\text{dev}}$ ($n{=}308$).}
\label{tab:per-class}
\end{table}

\begin{table}[ht]
\centering
\small
\setlength{\tabcolsep}{5pt}
\begin{tabular}{lcc}
\toprule
\textbf{Comparison} & \textbf{$\Delta$ F1} & \textbf{95\% CI} \\
\midrule
P1+P2 $\rightarrow$ P1+P3 & -0.1\% & [-5.9\%, 5.9\%] \\
P3 no filter $\rightarrow$ prefilter & 1.7\% & [-0.9\%, 4.1\%] \\
DeepSeek P1 $\rightarrow$ P1+P3 & 6.8\% & [-0.1\%, 13.2\%] \\
Llama P1 $\rightarrow$ P1+P3 & 15.2\% & [7.5\%, 23.1\%] \\
\bottomrule
\end{tabular}
\caption{\rev{Paired bootstrap confidence intervals} on $D_{\text{dev}}$ ($n{=}308$; 1,000 resamples).}
\label{tab:bootstrap}
\end{table}

\clearpage
\section{Embedding Space Analysis}
\label{app:embeddings}

To probe whether semantic similarity separates the response-clarity labels, we project 3{,}756 train+dev embeddings from Gemini Embedding 001 (3{,}072 dimensions; \citealp{google2026geminiembedding}) into two dimensions using UMAP \citep{mcinnes2018umap}. Figure~\ref{fig:umap} shows near-total class overlap, with a silhouette score of 0.001 indicating essentially no separation in this projection. All three classes intermix throughout the projection, and class centroids are separated by only 0.001--0.016 cosine distance in the projected space. \rev{This projection suggests that these embeddings do not cleanly separate response-clarity labels,} matching contrastive RAG ablations that did not improve classification. Evasion detection depends on pragmatic cues such as commitment strength, hedging patterns, and rhetorical structure that general-purpose text embeddings did not capture in this setting.

\begin{figure}[ht]
\centering
\includegraphics[width=\textwidth]{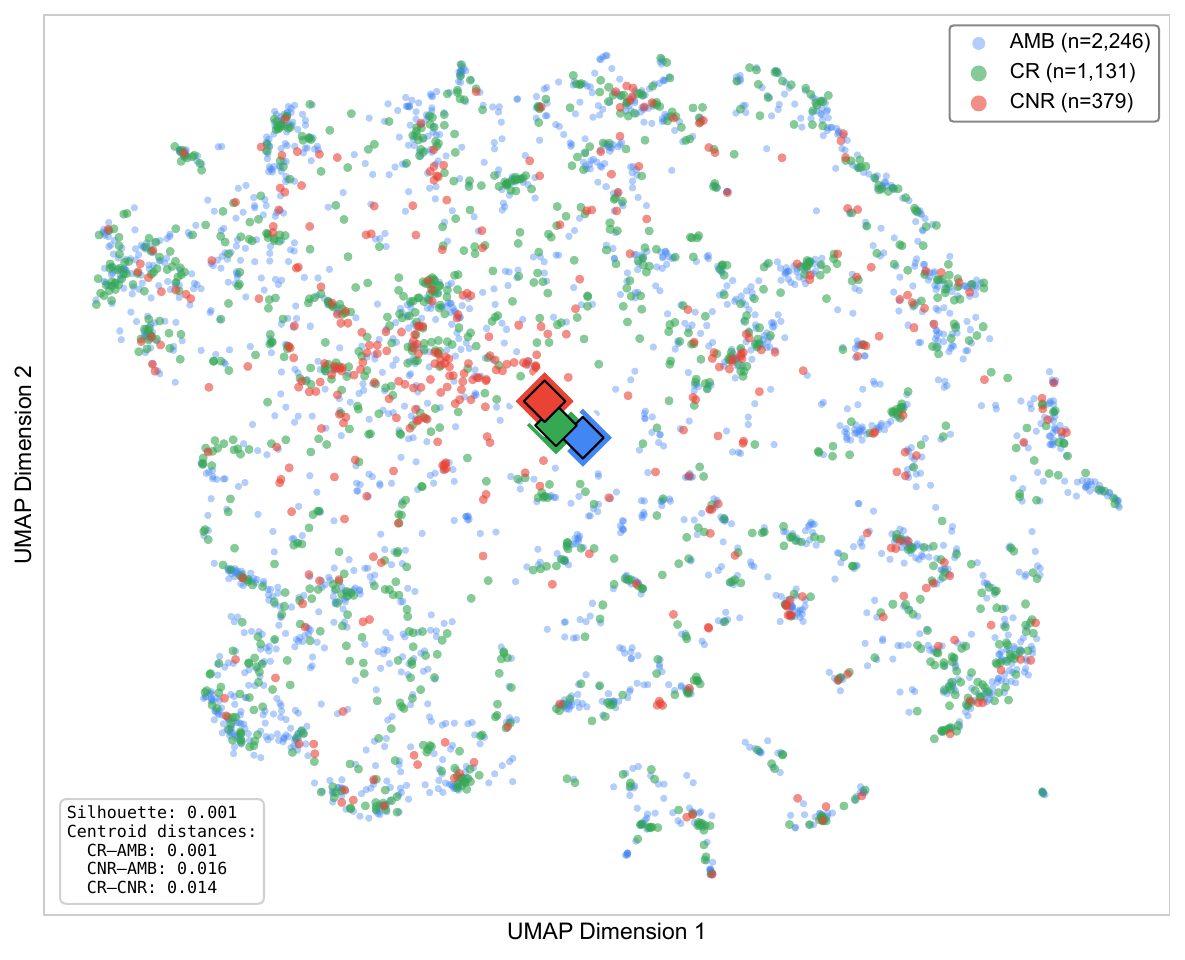}
\caption{UMAP projection of 3{,}756 train+dev embeddings (Gemini Embedding 001, 3{,}072 dimensions). Classes overlap almost entirely (silhouette = 0.001), suggesting that this embedding/projection does not cleanly separate response-clarity labels.}
\label{fig:umap}
\end{figure}

\clearpage
\section{Extended Qualitative Examples}
\label{app:examples}

Tables~\ref{tab:examples-correct} and~\ref{tab:examples-fail} show representative predictions with full reasoning traces. We select examples where verification \emph{changed} the initial prediction, demonstrating the pipeline's corrective behavior.

\begin{center}
\small
\captionof{table}{Verification saves: cases where Pass~2/3 corrected an initial error.}
\label{tab:examples-correct}
\vspace{0.5em}

\noindent\fbox{\parbox{0.96\textwidth}{%
\textbf{\cmark~Pass~3 upgrade: AMB $\rightarrow$ CR} \hfill \textit{Gold: CR} $\mid$ \textit{Final: CR}

\smallskip
\textit{Q: Are you committed to building the 700 miles of fence, actual fencing?}\\
A: ``\textbf{Yes}, we're going to do both, Joe. We're just going to make sure that we build it in a spot where it works\ldots''

\smallskip
\colorbox{bestrow}{\parbox{0.93\textwidth}{\textbf{Pass~1} (confidence 0.85 $\rightarrow$ AMB): ``The speaker begins with a direct `Yes,' which seems like a clear reply, but immediately qualifies the commitment by focusing on building fencing `in a spot where it works' rather than confirming the specific 700-mile target. The response employs deflection, shifting the focus from the specific mileage to a broader discussion about `smart borders,' guest worker programs, and terrain conditions.''}}

\smallskip
\colorbox{bestrow}{\parbox{0.93\textwidth}{\textbf{Pass~3} (upgrade $\rightarrow$ CR): ``The first substantive sentence begins with `Yes,' directly answering the question, and does not use prohibited preambles or immediate negations like `but' or `however'.''}}

\smallskip
{\footnotesize Pass~1 was distracted by post-answer qualifications; Pass~3 correctly anchored on the opening ``Yes.''}
}}

\bigskip

\noindent\fbox{\parbox{0.96\textwidth}{%
\textbf{\cmark~Pass~2a downgrade: CR $\rightarrow$ AMB} \hfill \textit{Gold: AMB} $\mid$ \textit{Final: AMB}

\smallskip
\textit{Q: Do you feel like you have been an effective leader when it comes to the economy?}\\
A: ``Well, first of all, wherever Republicans indicate an interest\ldots When I came into office, the U.S.\ economy had contracted by 9 percent\ldots Little over a year later, the economy was growing by 4 percent\ldots \textbf{It's not good enough}.''

\smallskip
\colorbox{bestrow}{\parbox{0.93\textwidth}{\textbf{Pass~1} (confidence 0.9 $\rightarrow$ CR): ``The speaker eventually provides a direct, substantive answer by citing specific economic metrics: `the U.S.\ economy had contracted by 9 percent\ldots a year later, the economy was growing by 4 percent.' This provides concrete numbers and dates to evaluate effectiveness, fitting the criteria for a Clear Reply.''}}

\smallskip
\colorbox{bestrow}{\parbox{0.93\textwidth}{\textbf{Pass~2a} (downgrade $\rightarrow$ AMB): ``The speaker presents evidence of economic growth but immediately qualifies it by saying it's `not good enough.' They never explicitly state whether they view themselves as an effective leader, leaving room for multiple interpretations about their self-assessment.''}}

\smallskip
\colorbox{bestrow}{\parbox{0.93\textwidth}{\textbf{Pass~3} (upgrade check $\rightarrow$ AMB, no change): ``The first substantive sentence starts with the preamble `Well, first of all' and pivots to discussing Republicans rather than directly answering the specific yes/no question about being an effective leader.''}}

\smallskip
{\footnotesize All three passes ran. Pass~1 was swayed by concrete metrics; Pass~2a caught the unanswered yes/no; Pass~3 confirmed the preamble signals evasion.}
}}

\bigskip

\noindent\fbox{\parbox{0.96\textwidth}{%
\textbf{\cmark~Pass~3 upgrade after parse failure: AMB $\rightarrow$ CR} \hfill \textit{Gold: CR} $\mid$ \textit{Final: CR}

\smallskip
\textit{Q: Are you still pursuing him?}\\
A: ``\textbf{Absolutely. No ands, ifs, or buts.} And in my judgment, it's just a matter of time, unless we stop looking\ldots''

\smallskip
\colorbox{bestrow}{\parbox{0.93\textwidth}{\textbf{Pass~1}: Parse failure (no structured output returned). Defaults to AMB with confidence 0.0.}}

\smallskip
\colorbox{bestrow}{\parbox{0.93\textwidth}{\textbf{Pass~3} (upgrade $\rightarrow$ CR): ``The first substantive sentence is `Absolutely,' which directly answers the question with a clear affirmative without preambles or immediate contrasting conjunctions.''}}

\smallskip
{\footnotesize The pipeline's fallback-to-AMB + mandatory verification recovered from a complete Pass~1 failure.}
}}

\bigskip

\noindent\fbox{\parbox{0.96\textwidth}{%
\textbf{\cmark~Pass~3 upgrade: AMB $\rightarrow$ CR} \hfill \textit{Gold: CR} $\mid$ \textit{Final: CR}

\smallskip
\textit{Q: Is the Vladimir Putin who said ``The United States is undermining global security'' the same Vladimir Putin whose soul you looked into and found to be trustworthy?}\\
A: ``I think \textbf{the person who I was referring to in 2001 is the same strong-willed person}. He is a person with whom I have had agreements and disagreements throughout the course of my Presidency\ldots''

\smallskip
\colorbox{bestrow}{\parbox{0.93\textwidth}{\textbf{Pass~1} (confidence 0.9 $\rightarrow$ AMB): ``The question asks if Putin is still the `trustworthy' person the speaker referenced in 2001. The speaker acknowledges Putin is the same `strong-willed' person but deflects to discussing policy disagreements on NATO and cooperation on non-proliferation. He avoids explicitly confirming or denying whether he still finds Putin `trustworthy.'\,''}}

\smallskip
\colorbox{bestrow}{\parbox{0.93\textwidth}{\textbf{Pass~3} (upgrade $\rightarrow$ CR): ``The first substantive sentence directly answers whether it's the same Putin by stating `I think the person who I was referring to in 2001 is the same strong-willed person.' This provides a clear position without contradictory qualifiers.''}}

\smallskip
{\footnotesize Pass~1 over-indexed on the ``trustworthy'' framing; Pass~3 recognized that identifying the same person \emph{is} the direct answer.}
}}

\end{center}

\clearpage

\begin{center}
\small
\captionof{table}{Verification failures: cases where Pass~2/3 introduced or failed to correct errors.}
\label{tab:examples-fail}
\vspace{0.5em}

\noindent\fbox{\parbox{0.96\textwidth}{%
\textbf{\xmark~Pass~3 over-upgrade: AMB $\rightarrow$ CR} \hfill \textit{Gold: AMB} $\mid$ \textit{Final: CR}

\smallskip
\textit{Q: Are you still willing to follow a path that seems to be in opposition to the will of the American people?}\\
A: ``\textbf{I am willing to follow a path that leads to victory}, and that's exactly why we're conducting the review\ldots Victory in Iraq is achievable. It hasn't happened nearly as quickly as I hoped\ldots''

\smallskip
\colorbox{highlight}{\parbox{0.93\textwidth}{\textbf{Pass~1} (confidence 0.85 $\rightarrow$ AMB): ``The answer uses deflection and reframing rather than providing a direct `yes' or `no' commitment to the question's premise. Instead of acknowledging or explicitly refusing the premise that his path is `in opposition to the will of the American people,' the speaker redefines the path as one that leads to `victory' and argues that the public actually supports staying.''}}

\smallskip
\colorbox{highlight}{\parbox{0.93\textwidth}{\textbf{Pass~3} (upgrade $\rightarrow$ CR): ``The first substantive sentence (`I am willing to follow a path that leads to victory') directly answers the question with a specific stance, contains no preambles, and is not immediately followed by a negating conjunction.''}}

\smallskip
{\footnotesize \textbf{Failure}: Pass~1 correctly identified the reframing (``a path that leads to victory'' $\neq$ the path asked about), but Pass~3's first-sentence heuristic overrode this nuanced judgment.}
}}

\bigskip

\noindent\fbox{\parbox{0.96\textwidth}{%
\textbf{\xmark~Pass~2a over-downgrade: CR $\rightarrow$ AMB} \hfill \textit{Gold: CR} $\mid$ \textit{Final: AMB}

\smallskip
\textit{Q: What have they achieved and what will they leave behind for their successors?}\\
A: ``Okay, I will start answering. Has it become better, or has it become worse?\ldots Let us remember the world on the brink of a nuclear disaster during the Caribbean crisis\ldots \textbf{We have created a reliable platform for cooperation}\ldots''

\smallskip
\colorbox{highlight}{\parbox{0.93\textwidth}{\textbf{Pass~1} (confidence 0.9 $\rightarrow$ CR): ``The speaker directly addresses the question by listing specific achievements in the bilateral relationship, such as shifting from enemies to partners and making progress in counterterrorism, non-proliferation, and fighting the drug threat.''}}

\smallskip
\colorbox{highlight}{\parbox{0.93\textwidth}{\textbf{Pass~2a} (downgrade $\rightarrow$ AMB): ``The answer is vague about specific achievements and doesn't clearly address what will be left behind for successors. While it mentions improved relations, it requires interpretation to understand the concrete nature and significance of these achievements.''}}

\smallskip
\colorbox{highlight}{\parbox{0.93\textwidth}{\textbf{Pass~3} (upgrade check $\rightarrow$ AMB, no change): ``The answer begins with the preamble `Okay, I will start answering,' followed by a rhetorical question (`Has it become better, or has it become worse?'). It fails to provide a direct answer in the first substantive sentence.''}}

\smallskip
{\footnotesize \textbf{Failure}: All three passes ran. Pass~1 correctly recognized substantive policy content; Pass~2a and~3 both penalized the rhetorical register despite the answer containing concrete achievements.}
}}

\bigskip

\noindent\fbox{\parbox{0.96\textwidth}{%
\textbf{\xmark~Pass~3 over-upgrade: CNR $\rightarrow$ CR} \hfill \textit{Gold: CNR} $\mid$ \textit{Final: CR}

\smallskip
\textit{Q: Why wouldn't you be frustrated, sir, about what's happening?}\\
A: ``\textbf{I'm not}---I do remember the meeting; I don't remember being surprised. I'm not sure what they meant by that.''

\smallskip
\colorbox{highlight}{\parbox{0.93\textwidth}{\textbf{Pass~1} (confidence 0.9 $\rightarrow$ AMB): ``The question asks for a reason why the speaker is not frustrated. The answer provides information about memory of a meeting and lack of surprise, but fails to address the specific emotional state of frustration or the reasoning behind it.''}}

\smallskip
\colorbox{highlight}{\parbox{0.93\textwidth}{\textbf{Pass~3} (upgrade $\rightarrow$ CR): ``The first substantive sentence (`I'm not---I do remember the meeting; I don't remember being surprised') directly denies the frustration implied in the question and provides a specific factual stance on the meeting without preambles or negating conjunctions.''}}

\smallskip
{\footnotesize \textbf{Failure}: The truncated ``I'm not---'' is a false start, not a denial. The speaker never engages with the frustration question, instead pivoting to meeting recall. This is a refusal by redirection, which neither pass detected.}
}}

\bigskip

\noindent\fbox{\parbox{0.96\textwidth}{%
\textbf{\xmark~Pass~3 over-upgrade: AMB $\rightarrow$ CR} \hfill \textit{Gold: AMB} $\mid$ \textit{Final: CR}

\smallskip
\textit{Q: Has the scandal damaged Hastert's credibility and effectiveness in maintaining party control in the midterm elections?}\\
A: ``\textbf{No}, I think the Speaker's strong statements have made it clear to not only the party members but to the country that he wants to find out the facts\ldots''

\smallskip
\colorbox{highlight}{\parbox{0.93\textwidth}{\textbf{Pass~1} (confidence 0.9 $\rightarrow$ AMB): ``The speaker begins with a direct `No,' but fails to substantively address the specific impact on `maintaining party control in the midterm elections.' Instead, the response deflects to a defense of the Speaker's statements, general condemnation of the behavior, and a counter-accusation.''}}

\smallskip
\colorbox{highlight}{\parbox{0.93\textwidth}{\textbf{Pass~3} (upgrade $\rightarrow$ CR): ``The first substantive sentence begins with `No,' which is a direct answer to the question, and it is not followed by immediate qualifiers or negations like `but' or `however'.''}}

\smallskip
{\footnotesize \textbf{Failure}: Pass~1 correctly noted the ``No'' is followed by topic deflection rather than substantive engagement with the midterm elections premise. Pass~3's first-sentence rule cannot distinguish a genuine ``No'' from a dismissive one.}
}}

\end{center}

\fi

\end{document}